\documentclass[10pt, a4paper, onecolumn]{article}

\usepackage{arxiv}
\usepackage[toc]{appendix}
\usepackage[utf8]{inputenc} 
\usepackage[T1]{fontenc}    
\usepackage{hyperref}       
\usepackage{url}            
\usepackage{booktabs}       
\usepackage{amsfonts}       
\usepackage{nicefrac}       
\usepackage{microtype}      
\usepackage{lipsum}
\usepackage{amsmath}
\usepackage{cite}
\usepackage{amssymb}
\usepackage{verbatim}
\usepackage{enumerate}
\usepackage{mdwlist} 
\usepackage{color}

\usepackage{algpseudocode}
\usepackage{algorithmicx}
\usepackage{lineno}
\usepackage{framed,multirow}
\usepackage[final]{pdfpages}
\usepackage{latexsym}
\usepackage{wrapfig}

\usepackage{url}
\usepackage[table]{xcolor}
\usepackage{xcolor}
\usepackage{graphicx}
\usepackage{subfigure}
\usepackage{verbatim}
\usepackage{bm}
\usepackage{authblk}
\usepackage{tikz}
\usepackage{amsmath}
\usepackage[linesnumbered,ruled]{algorithm2e}
\usepackage[space]{grffile}
\usetikzlibrary{shapes,arrows}

\title{VORTEX: Challenging CNNs at Texture Recognition by using Vision Transformers with Orderless and Randomized Token Encodings 
}

\author[1]{Leonardo Scabini}
\author[1]{Kallil M. Zielinski}
\author[2]{Emir Konuk}
\author[3]{Ricardo T. Fares}
\author[3]{Lucas C. Ribas}
\author[2]{Kevin Smith}
\author[1]{Odemir M. Bruno}

\affil[1]{\small{S\~{a}o Carlos Institute of Physics, University of S\~{a}o Paulo, São Carlos - SP, Brazil}}
\affil[2]{\small{KTH Royal Institute of Technology, and Science for Life Laboratory, Stockholm, Sweden}}
\affil[3]{\small{São Paulo State University, Institute of Biosciences, Humanities and Exact Sciences, São José do Rio Preto - SP, Brazil}}

\begin{document}
\maketitle
\begin{abstract}
Texture recognition has recently been dominated by ImageNet-pre-trained deep Convolutional Neural Networks (CNNs), with specialized modifications and feature engineering required to achieve state-of-the-art (SOTA) performance. However, although Vision Transformers (ViTs) were introduced a few years ago, little is known about their texture recognition ability. Therefore, in this work, we introduce VORTEX (ViTs with Orderless and Randomized Token Encodings for Texture Recognition), a novel method that enables the effective use of ViTs for texture analysis. VORTEX extracts multi-depth token embeddings from pre-trained ViT backbones and employs a lightweight module to aggregate hierarchical features and perform orderless encoding, obtaining a better image representation for texture recognition tasks. This approach allows seamless integration with any ViT with the common transformer architecture. Moreover, no fine-tuning of the backbone is performed, since they are used only as frozen feature extractors, and the features are fed to a linear SVM. We evaluate VORTEX on nine diverse texture datasets, demonstrating its ability to achieve or surpass SOTA performance in a variety of texture analysis scenarios. By bridging the gap between texture recognition with CNNs and transformer-based architectures, VORTEX paves the way for adopting emerging transformer foundation models. Furthermore, VORTEX demonstrates robust computational efficiency when coupled with ViT backbones compared to CNNs with similar costs. The method implementation and experimental scripts are publicly available in our online repository~\footnote{\label{github}\url{https://github.com/scabini/VORTEX}}.
\end{abstract}

\section{Introduction}

Analyzing visual textures in digital images has been an important research area in Computer Vision (CV) for several decades. While CV is a vast field dealing with all aspects of visual recognition (shapes, objects, etc), focusing solely on textures allows us to solve a range of specific and fine-grained problems. In this sense, texture recognition methods have been successfully employed in various applications
~\cite{ghalati2021texture}. From hand-engineered approaches to deep-learning-based models, the texture recognition state-of-the-art (SOTA) is constantly evolving throughout the years~\cite{liu2019bow}. 

Lately, several deep-learning-based models have been proposed~\cite{zhai2020dsrnet,condori2021rankgp3mcnn,chen2021classnet,yang2022,lyra2022multilayerfv,scabini2023radam,liu2024dual,qiu2024hrnet,florindo2024fractal} specially due to the large-scale pre-training and transfer-learning of deep Convolutional Neural Networks (CNNs)~\cite{lecun1998gradient,krizhevsky2017imagenet}. However, these pre-trained CNNs need additional architecture engineering or fine-tuning to perform above other texture recognition techniques. This approach is necessary because while off-the-shelf pre-trained CNN features can be ported to texture recognition with minimal modifications, the obtained results may fall short compared to hand-engineered methods~\cite{scabini2020spatio}. Therefore, additional feature engineering is needed to improve foundation models (backbones) pre-trained on natural images by addressing their ability to distinguish textures. This is usually achieved by improving feature pooling through more sophisticated orderless operations, feature aggregation using layers from different depths, and adding extra learnable layers/modules.

On the other hand, Vision Transformers (ViT)~\cite{dosovitskiy2020image} have been recently dominating many CV areas~\cite{khan2022transformers}. They represent a paradigm shift in architectural design from CNNs, proposing a less bio-inspired structure by focusing more on parallelizable operations and larger receptive fields, better capturing longer-distance relationships among visual cues. However, little is known about their texture recognition ability. While it has been shown that they may compete with CNNs~\cite{scabini2024comparative} when using their off-the-shelf features, ViTs may struggle to surpass even hand-engineered methods in some specific cases. This behavior highlights the need for new methods of architecture adaptations and feature engineering to improve foundation ViT backbones on texture recognition. For instance, Liu et al.~\cite{liu2024dual} employ transformers for texture recognition, but they are used only for feature engineering over CNN backbones rather than ViT backbones.

\begin{figure}[!htb]
  \centering
  
  \includegraphics[width=0.95\linewidth]{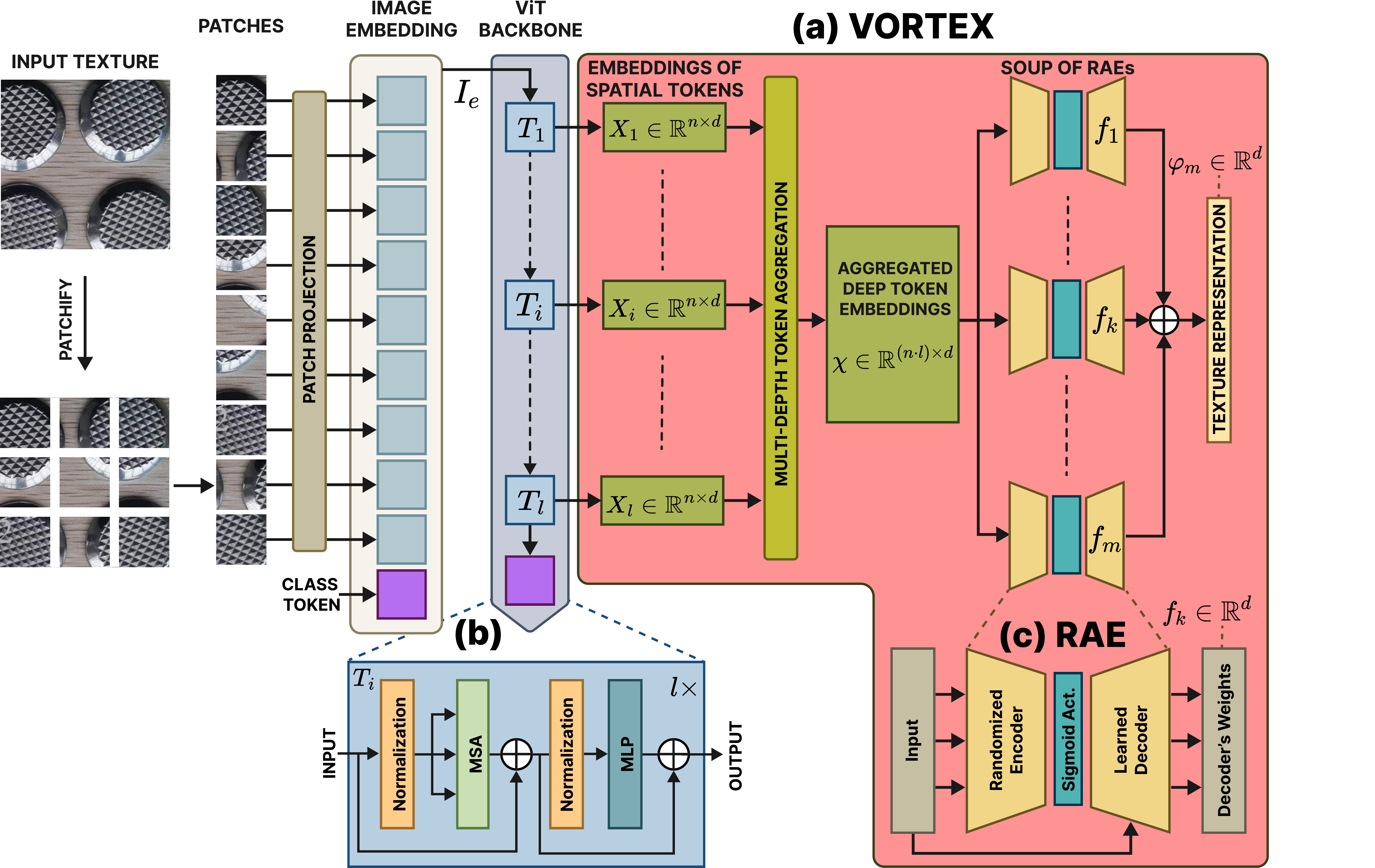}
  
  \caption{Illustration of the proposed method for feature engineering with pre-trained ViTs, VORTEX (a), used in this example with the vanilla ViT architecture (b) to produce a new image representation $\varphi_m$. The structure of the RAE is shown in (c), which is an analytically-solved 1-layer auto-encoder, where we use its decoder weights as the representation (souped for $m$ RAEs).}
  \label{fig:method}
\end{figure}

In this work, we propose a new texture recognition method that takes advantage of several foundation ViTs. We achieve that by building a new module, we named VORTEX, based on pre-trained \textbf{V}iTs with \textbf{O}rderless and \textbf{R}andomized \textbf{T}oken \textbf{E}ncodings to extract te\textbf{X}ture features. Our method is the first to perform feature engineering on ViT backbones focusing on general texture recognition, and it works with any backbone with the common transformer architecture. Therefore, we evaluate VORTEX using a variety of backbones with different sizes and pre-trainings. Moreover, no fine-tuning is performed as VORTEX does not add new trainable parameters to the ViT backbone, as it only performs feature extraction with frozen backbones. The overall structure of VORTEX is illustrated in Figure~\ref{fig:method} (a), where an image representation is obtained given an input texture image and a ViT backbone. The representations are then used to train a simple linear classifier, making it also more feasible and efficient on small datasets. We evaluate the proposed method using nine texture datasets and compare its results with CNN and ViT techniques from the literature. The VORTEX module code and scripts for experimentation are available in our public GitHub repository~\ref{github}.

\section{Background}

This section presents the fundamental principles of texture analysis and ViTs, also pointing out the gaps in these fields that pave the way to the proposed VORTEX method.

\subsection{Texture Analysis}




Since the emergence of the CV field, much effort has been devoted to enabling computer systems to perform actions such as object detection and material recognition, which are crucial skills in several real-world applications. To achieve that, one of the most important approaches is the recognition of textures. Texture is a fundamental and inherent property of surfaces and objects. In digital images, it can be characterized by the spatial arrangement of the pixels and/or the local relationships of a pixel with its surrounding neighborhood \cite{haralick1973textural}.
In this context, the texture analysis field has steadily evolved towards more robust and discriminative texture representations. 
Initially, several hand-engineered approaches have been proposed relying on pixel statistics \cite{haralick1973textural}, filters and wavelets \cite{manjunath1996texture}, local-based patterns~\cite{ojala2002multiresolution}, among others~\cite{liu2019bow}. 


Due to the surprising CV results obtained by pure learning-based techniques such as CNNs~\cite{lecun1998gradient,krizhevsky2017imagenet}, there has been a shift in the last decade towards deep learning techniques for texture analysis. This change was also motivated by hand-engineering methods struggling in more complex scenarios~\cite{cimpoi2014describing} such as internet images or when multiple objects and backgrounds are present. In this context, the rapid advancements of deep learning have created a common sense that they are universal solvers for CV tasks. However, off-the-shelf deep CNNs may fail to achieve SOTA performance in texture recognition tasks compared to hand-engineered approaches \cite{scabini2020spatio,scabini2024comparative}. This limitation arises because many aspects of texture remain challenging to model, even in controlled imaging scenarios. Some examples include variations in lightning conditions, high texture variability (such as the same materials and surfaces under different conditions), and high texture similarities (in fine-grained tasks such as distinguishing different grades of the same material), among others.

To address these challenges, deep learning-based methods for texture recognition have focused on customizing pre-trained CNNs through feature-engineering and architecture modifications~\cite{zhai2020dsrnet,condori2021rankgp3mcnn,chen2021classnet,yang2022,lyra2022multilayerfv,scabini2023radam,liu2024dual,qiu2024hrnet,florindo2024fractal}. These approaches usually focus on enhancing pre-trained CNN feature extraction and/or end-to-end fine-tuning strategies. Among the feature extraction operations, most methods adopt orderless poolings or encodings over feature/activation maps obtained from neural network layers, as prior research has demonstrated~\cite{cimpoi2014describing} that enforcing spatial order can limit the texture recognition ability. More recently, the proposal of ViTs~\cite{dosovitskiy2020image} has been leading to performance improvements on various CV tasks~\cite{khan2022transformers}, also holding promise for texture recognition~\cite{scabini2024comparative}. This architecture excels on capturing longer-range correlations among visual cues, compared to CNNs. However, little is known about how ViTs behave compared to CNNs modified for texture analysis and whether such feature-engineering and architecture changes could improve the ability of ViTs to recognize textures.

\subsection{Vision Transformers}

Transformers have revolutionized many fields in artificial intelligence, originally demonstrating unprecedented success in natural language processing (NLP) tasks ~\cite{vaswani2017attention}. By leveraging a highly parallelizable architecture, transformers exhibit faster training and superior performance in machine translation, outperforming prior sequence-to-sequence NLP approaches. The original transformer design is composed of encoders and decoders, containing two main structures: Multiheaded Self-Attention (MSA) and a Feed Forward Network (FFN).

Let us first consider a set of input tokens, such as word embeddings combined into a matrix $I$. These tokens are transformed via linear projections into three matrices:  $Q_i = IW_{Q_i}$ (queries), $K_i = IW_{K_i}$ (keys) and $V_i = IW_{V_i}$ (values), where $\{W_{Q_i}, W_{K_i}, W_{V_i}\}$ represent trainable weight matrices. The self-attention for each attention head $i$ is then computed as: 
 \begin{equation} 
 \label{eq:self-attention} 
 \text{Attention}(Q_i, K_i, V_i) = \text{softmax}\Big(\frac{Q_i K_i^{T}}{\sqrt{d}}\Big) V_i\,,
 \end{equation}
 where $d$ denotes the dimensionality of each projection, and the softmax function is applied row-wise. Multiple heads are then stacked in parallel:
\begin{equation} 
\label{eq:MSA} 
\text{MSA}(Q, K, V) = \big[\text{Attention}(Q_1, K_1, V_1) || \cdots || \text{Attention}(Q_s, K_s, V_s)\big] W_o\,,
\end{equation}
where $||$ indicates concatenation, $s$ is the number of attention heads, and $W_o$ is a learnable projection matrix applied after concatenation. A simple two-layer MLP is used as the FFN, and both the MSA and FFN are surrounded by skip connections and layer normalization~\cite{vaswani2017attention}. A transformer network is composed of stacked ``transformer blocks'', as illustrated in Figure~\ref{fig:method} (b), each containing independent MSA and FFN layers.

Adapting transformers for CV tasks, the Vision Transformer (ViT) ~\cite{dosovitskiy2020image} introduces a novel approach by treating images as sequences of 2D patches. Given an image $I \in \mathbb{R}^{w\times h \times 3}$, with width $w$ and height $h$, it is is divided into $n = \frac{wh}{p^2}$ patches, where $p\times p$ is the patch size. Each patch is flattened into a vector of size $3p^2$ for 3-channel RGB images, forming a sequence of embeddings $I_p \in \mathbb{R}^{n \times (3p^2)}$. A learnable linear projection $E \in \mathbb{R}^{(3p^2) \times d}$ is then applied to produce patch embeddings $I_e \in \mathbb{R}^{n \times d}$, augmented with positional encoders $E_{pos}$ to retain spatial information. The resulting embeddings are processed by a series of transformer blocks, and a classification head is attached to a learnable class token (CLS). Figure~\ref{fig:method} illustrates some of the main steps in a ViT model.

ViTs have demonstrated SOTA performance across various CV tasks~\cite{dosovitskiy2020image,khan2022transformers}. However, while their success in general CV tasks is well-documented, their application to texture analysis remains underexplored. For instance, Zhang et.al.~\cite{Zhang2021} applied ViTs to steel texture recognition, achieving superior accuracy compared to CNNs and traditional machine learning approaches. Soleymani et al.~\cite{soleymani2021} demonstrated the efficacy of ViTs in building material classification, particularly in unbalanced datasets. A wide evaluation of pre-trained ViT architectures is depicted by Scabini et al.~\cite{scabini2024comparative}. These studies show the potential of ViTs in texture analysis, but they focus on applying existing pre-trained architectures without exploring architectural modifications tailored specifically to texture recognition.

In the realm of CNNs, it is well-established that customizing architectures and aggregating features from intermediate layers can substantially enhance texture recognition~\cite{zhai2020dsrnet,condori2021rankgp3mcnn,chen2021classnet,yang2022,lyra2022multilayerfv,scabini2023radam,liu2024dual,qiu2024hrnet,florindo2024fractal}. The same principle could be extended to ViTs, as highlighted by recent works investigating hierarchical transformers~\cite{wang2021pyramid, Liu2021SwinTH}. 
On the other hand, some works have explored feature aggregation in ViTs for 
remote sensing~\cite{mao2022mfatnet}, and image super-resolution~\cite{yoo2023enriched}. Another work~\cite{liu2024dual} also proposes transformers for feature engineering over CNN backbones. However, none of these works focus on enhancing the textural features of pre-trained ViTs and the performance impacts on known texture recognition benchmarks. Therefore, leveraging the strengths of ViTs could pave the way for building better texture recognition methods, potentially breaking the hegemony of CNNs in this field.

\section{Proposed Method: VORTEX}

Considering the literature gaps in texture recognition and ViTs, we propose a new method named VORTEX. In this section, we describe the proposed method in depth, showing how to couple it with different ViT backbones, extracting improved image descriptors from them that are coupled with linear classifiers for texture recognition.

\subsection{Multi-depth Token Aggregation}

The first step of most feature-engineering/learning methods for texture analysis with deep CNN backbones consists of extracting features, or activations, at multiple stages of the backbone. This approach is beneficial since it allows extracting more low-level features from earlier layers, which usually focus more on less complex visual cues such as local texture patterns. In a ViT, the feature structure is particularly different from CNNs since intermediate activations represent token embeddings. Nevertheless, these embeddings encode visual patterns at varying hierarchical levels, corresponding to different regions of the image (patches/tokens). To extract these embeddings, in practice, consider the matrix $I_{e}$ as the input embeddings of a ViT, obtained given an input image $I$, which may include positional embeddings, convolutional embeddings, or other design choices. By feeding $I_{e}$ into a transformer encoder $B$ as in the original ViT model~\cite{dosovitskiy2020image}, at any transformer block $T_i$ we can obtain a 2-dimensional tensor (ignoring the batch dimension, for simplicity) $X_i \in \mathbb{R}^{n \times d}$ representing the spatial tokens, where $n$ is the number of tokens and $d$ the hidden dimension of the architecture. The CLS token is discarded in this case. 

\begin{figure}[!htb]
  \centering
  \subfigure[Input textures (fabric, leaf, metal object, and wrinkled surface).]{ 
   \includegraphics[width=0.15\linewidth]{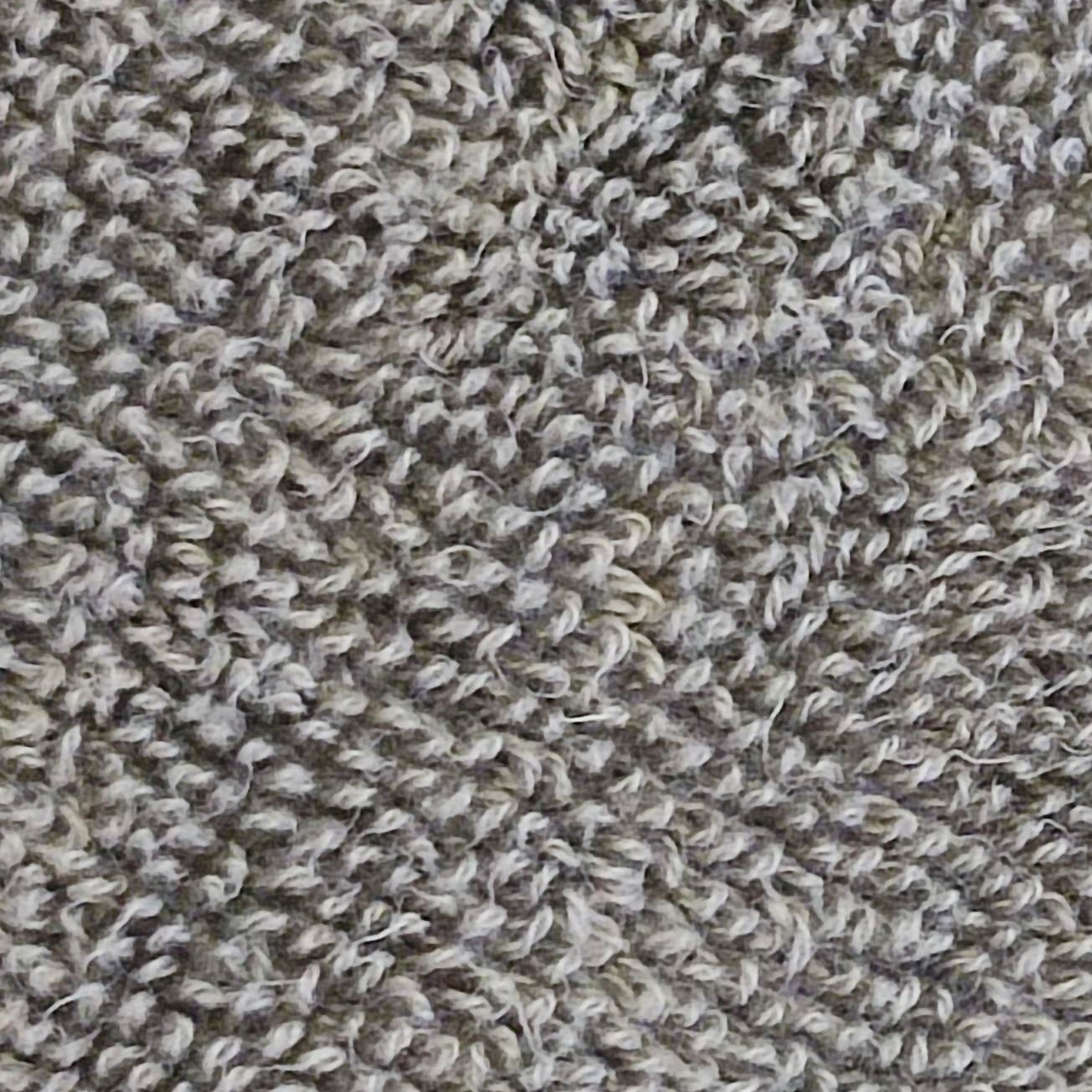}
   \includegraphics[width=0.15\linewidth]{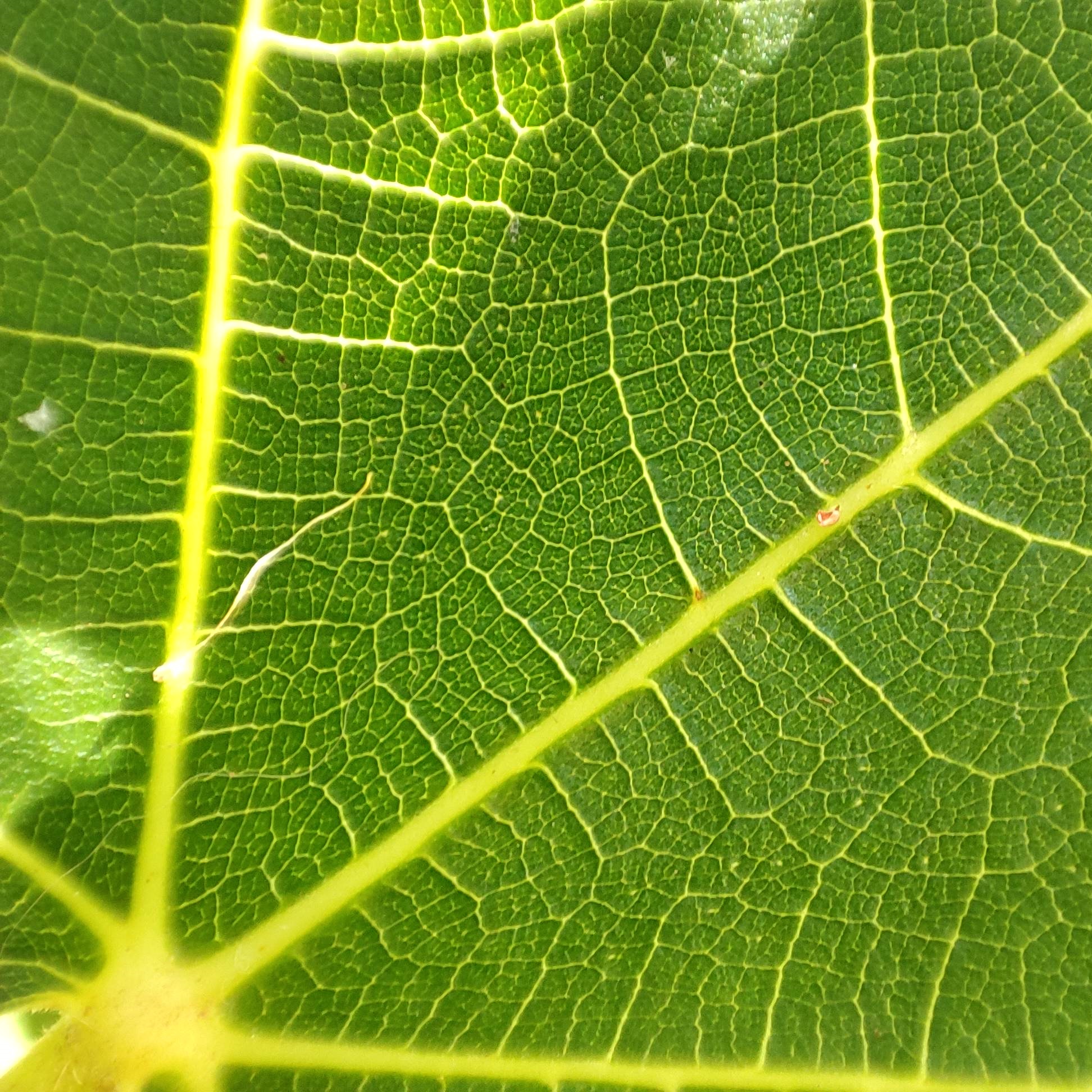}
   \includegraphics[width=0.15\linewidth]{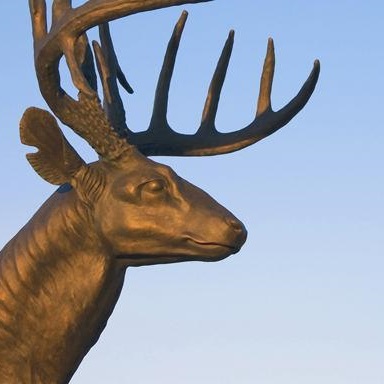}
   \includegraphics[width=0.15\linewidth]{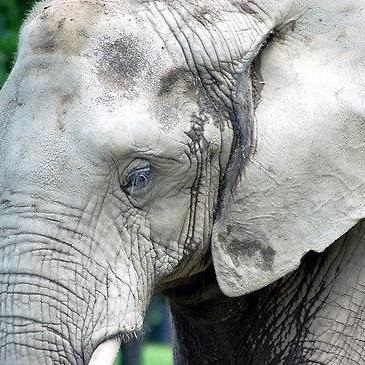}
   }
   \\
  \subfigure[ViT-B/16 attention scores at spatial tokens.]{ 
   \includegraphics[width=0.9\linewidth]{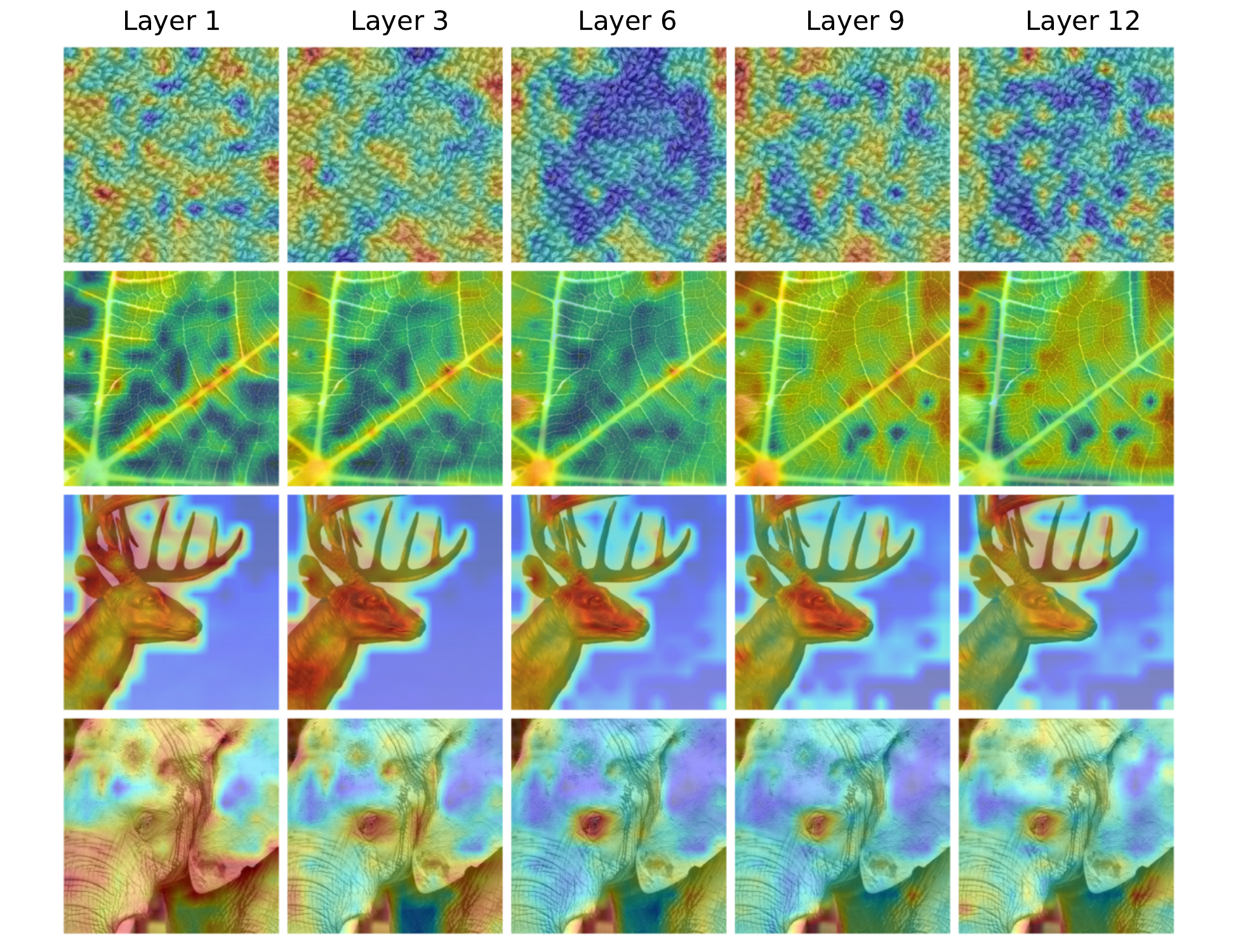}
     }

  \caption{Attention scores computed by averaging the attention heads and self-similarity scores among spatial tokens at different layers of ViT-B/16 (IN-21k pre-training)~\cite{dosovitskiy2020image}. Warmer colors represent higher attention.}
  \label{fig:attention}
\end{figure}

To illustrate the function of spatial tokens at different ViT layers, we extract attention scores at these layers for different texture samples. For this analysis, we consider the ViT-B/16~\cite{dosovitskiy2020image} (IN-21k pre-training) model, and the scores are calculated by obtaining the softmax output after the multi-head self-attention operation, averaging the attention heads, and averaging the scores among the spatial tokens. Figure~\ref{fig:attention} shows the obtained scores for layers 1, 3, 6, 9, and 12 of ViT-B/16 using four different texture images. It is possible to observe that the model progressively focuses on different visual cues. For pure-texture images (i.e., no object, background, etc.), such as the fabric and leaf samples, the attention in shallow layers is more sparse and global. At the last layers, on the other hand, the model tries to focus on specific regions, reflecting the object detection bias due to its natural vision pre-training. Similarly, when there are complex objects or scenes with backgrounds, such as the metal object and the elephant with the wrinkled texture, the attention ranges from more general and wider areas at shallow layers to more specific and complex patterns at deeper layers, such as eyes, tusks, and antlers. This behavior is expected (and necessary) for object detection and recognition in general vision tasks. Moreover, the deep features are important to ensure robustness under complex scenarios, and they may be enough in some texture analysis cases, e.g., the eyes of the animals may contain enough cues about the surface material or pattern. Nevertheless, the most important regions for distinguishing textures are the larger surfaces of the objects, which are crucial to distinguish their material (e.g. fabric, leaf, metal) or texture attribute (wrinkled). Therefore, features from shallow layers are also beneficial for tasks that rely on texture recognition.

To improve the texture recognition capabilities of ViTs, we propose to extract the spatial token embeddings at all layers, which encode the hierarchical feature processing of the ViT backbone. To aggregate the information from the different tokens at different depths, concatenating $X_i \forall i \in \{1, \dots, l\}$ gives a matrix with $ln$ lines and $d$ columns:
\begin{equation}
    \chi' = [X_1 ; \dots ; X_l ] \in \mathbb{R}^{ln \times d}\,,
\end{equation}
 where $[.;.]$ denotes a concatenation along the first dimension (lines). Additionally, an $l_2$-normalization is applied to balance feature magnitudes and improve numerical stability for downstream tasks, resulting in $\chi \in \mathbb{R}^{ln \times d}$:
\begin{equation}
\chi_{ij} = \frac{\chi'_{ij}}{\|\chi'_j\|_2}, \quad \|\chi'_j\|_2 = \sqrt{\sum_{i=1}^{ln} {\chi'_{ij}}^2}, \quad \forall j \in \{1, 2, \dots, d\}.
\end{equation}

 We refer to $\chi$ as the aggregated deep token embeddings, which are treated as a transformer-based representation of the input image $I$, containing multi-depth hierarchical features from the pre-trained ViT. These steps are illustrated in Figure~\ref{fig:method}, which shows the overall structure of the proposed method.

\subsection{Orderless Randomized Token Encoding}

Although $\chi$ stores different levels of hierarchical features across tokens, it needs further summarization to compose a one-dimensional image representation. Moreover, the aggregation step preserves the flattened token order and may result in features with more sensitivity to spatial structures. While this benefits object recognition, it is suboptimal for texture analysis. Unlike objects, textures are characterized by repeating stochastic or regular patterns rather than fixed spatial arrangements or geometric shapes. Therefore, an orderless operation is required to encode token embeddings in a way that captures the overall texture features rather than their absolute positions, making the texture representation more robust to variations in texture rotation, scale, and local deformations.

To achieve that, we propose an approach based on Randomized Autoencoders (RAEs)~\cite{RNN-AE}, which is a type of Randomized Neural Network. Therefore, $\chi$ is considered as the training input for RAEs, where each training sample is a token, and each column is a feature from the $d$-dimensional ViT embedding space (or hidden size). Therefore, the training matrix $\chi \in \mathbb{R}^{(ln) \times d}$, with $ln$ training samples with $d$ features, is built by organizing the aggregated tokens and embedding dimensions into lines and columns, respectively. 

The RAE is a 1-hidden-layer randomized neural network, as illustrated in Figure~\ref{fig:method} (c). Its first step is to project the input $\chi$ using a random fully-connected layer with weights $W_k \in \mathbb{R}^{d \times q}$, followed by a sigmoid nonlinearity. The weights are generated using the Linear Congruential Generator (LCG) for simplicity and better reproducibility, followed by standardization (zero-centered, unit variance) and orthogonalization~\cite{saxe2013exact}. As for the LCG parameters, we use $a=75$, $b=74$, and $c=2^{16}+1$, starting with $x=0$, a classical configuration according to the ZX81 computer from 1981. Here $k$ works like a seed for random sampling, denoting a starting index inside the LCG space generated with the given configuration. These configurations were chosen following previous works~\cite{ribas2018fusion,RNN-AE,scabini2023radam}. The forward pass of the encoder $g_k \in \mathbb{R}^{ln \times q}$ for all samples is then obtained as:
\begin{equation}
    g_k = \phi(\chi W_k)\,,  
\end{equation}
and the training of the decoder weights $f_k \in \mathbb{R}^{q \times d}$ is obtained as the least-squares solution:
\begin{equation}
    f_k =  g_k^{T}(g_kg_k^{T})^{-1} \chi\,.
\end{equation}

In this sense, given one input image, we train an RAE over the aggregated deep token embeddings $\chi$ obtained after the forward pass of the backbone. The main idea of employing an individual randomized neural network for each image is to use the output weights as a representation. This is possible since the RAE will always have the same fixed random encoder, defined by the LCG, while the decoder is learned for each image. To maintain feature dimensionality, we consider a single hidden neuron $q=1$. In this sense, the resulting decoder weights are represented by:
\begin{equation}
    f_k = (\nu_1, \ldots, \nu_d)\,, 
\end{equation}
where $\nu_i$ represents the connection weight between the single hidden neuron and the output $i$, corresponding to feature $i \in \{ 1, \dots, d \}$. It is important to notice that since these operations are invariant to the input sample order (row order), this process becomes an orderless encoding of all $ln$ tokens, from all transformer blocks, into the learned $f_k$ representation. The overall structure of the RAE is illustrated in Figure~\ref{fig:method} (c). 

A single RAE may have limited encoding capacity. Therefore, we use a model ``soup'' built by combining the weights of $m$ parallel RAEs. To achieve that, we use different LCG seeds, thus generating encoders with different random weights. The combination (``soup'') is then performed by summing the learned decoder weights:
\begin{equation}
    \varphi_m = (\sum_{k=1}^m f_k(\nu_1) ,\ldots, \sum_{k=1}^m f_k(\nu_d))\,.
\end{equation}

The feature vector $\varphi_m$ represents the final 1-dimensional image representation of the input texture, given the number of encoders in the soup as the parameter $m$. It combines the capacity of different RAEs, encoding token embeddings from all layers, ranging from the shallow features to the more complex deep ones, while being invariant to the token/patch order. We name this approach \textbf{V}iTs with \textbf{O}rderless and \textbf{R}andomized \textbf{T}oken \textbf{E}ncodings to extract te\textbf{X}ture (VORTEX) features. The VORTEX features can then be used as input to an output head or other models to perform different pattern recognition tasks. Here, we focus on image classification, so we consider linear classifiers as the last layer of the model. Moreover, since our focus is on feature extraction, the backbone is frozen and no backpropagation is performed. Therefore, VORTEX performs off-the-shelf feature extraction with the ViT backbone, and a separate linear classifier is trained with the obtained features.

\section{Experiments and Discussion}

This section provides all technical details and presents the experimental results to validate the proposed VORTEX approach, along with its comparisons to SOTA techniques.

\subsection{Experimental Setup} 

\begin{figure}[!htb]
    \centering
        \subfigure[Outex10]{\includegraphics[width=0.3\linewidth]{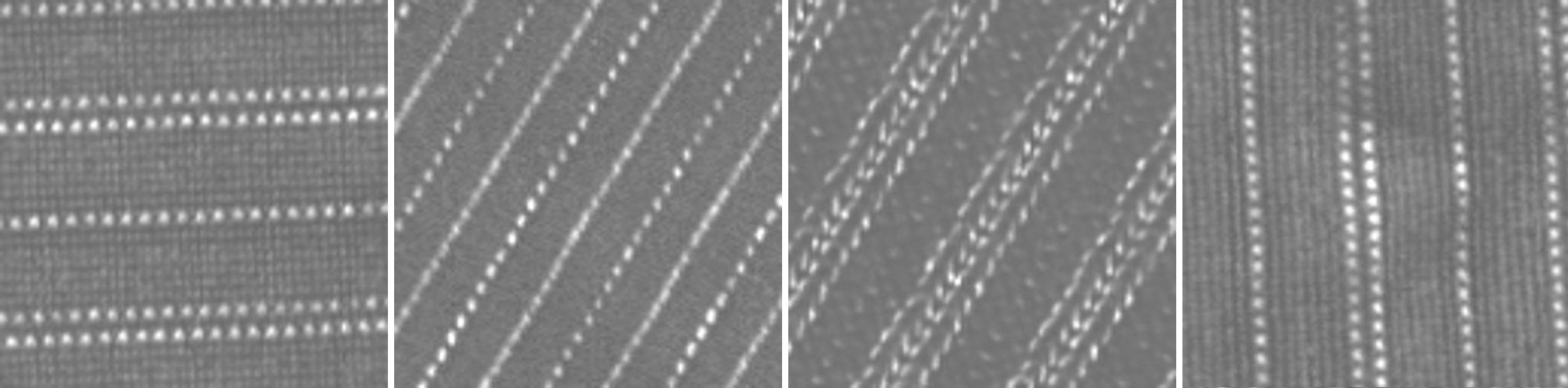}} \hspace{0.2cm} \subfigure[Outex12]{\includegraphics[width=0.3\linewidth]{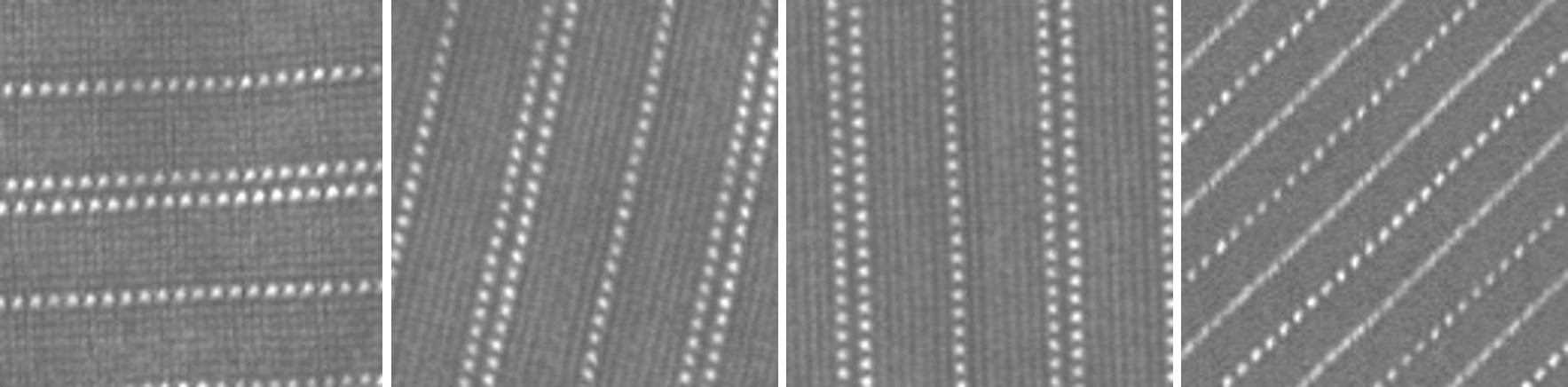}} \hspace{0.2cm}  \subfigure[Outex13]{\includegraphics[width=0.3\linewidth]{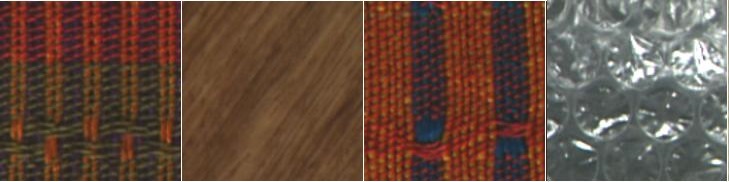}} \\

        \subfigure[Outex14]{\includegraphics[width=0.3\linewidth]{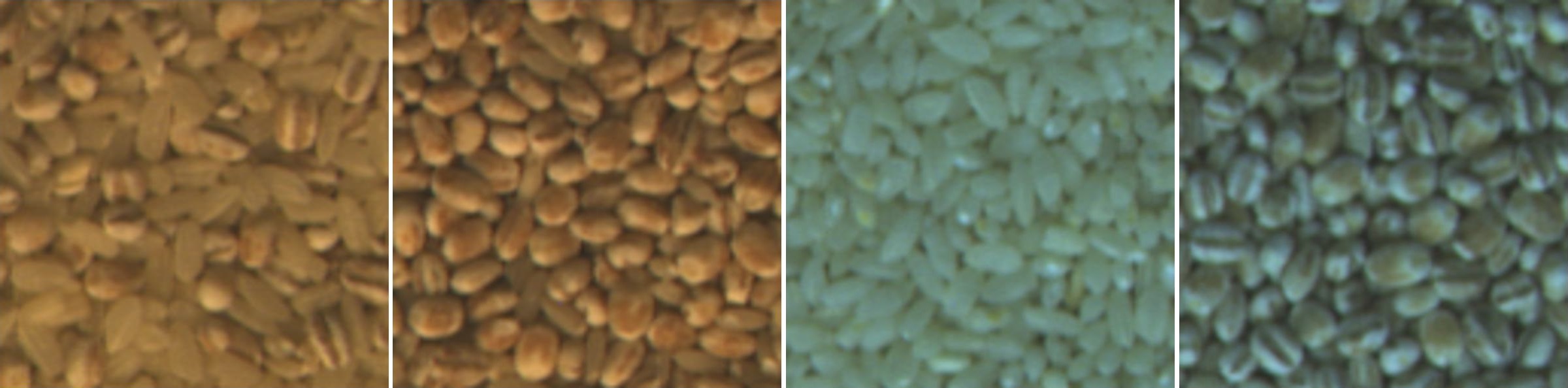}} \hspace{0.2cm} \subfigure[DTD]{\includegraphics[width=0.3\linewidth]{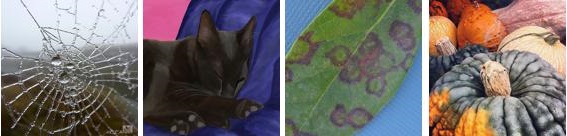}} \hspace{0.2cm} \subfigure[FMD]{\includegraphics[width=0.3\linewidth]{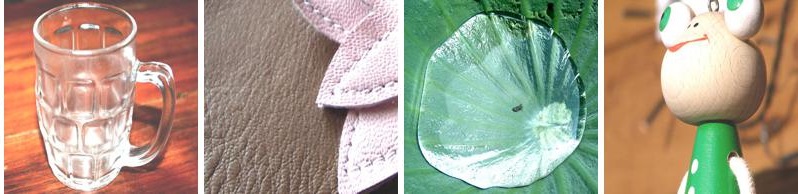}} \\

        \subfigure[KTH-2-b]{\includegraphics[width=0.3\linewidth]{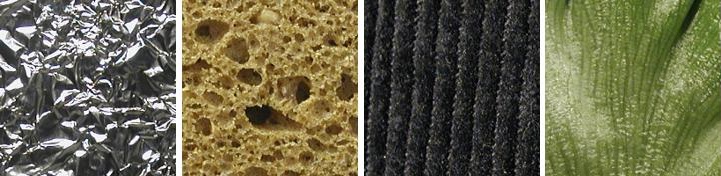}} \hspace{0.2cm}\subfigure[GTOS]{\includegraphics[width=0.3\linewidth]{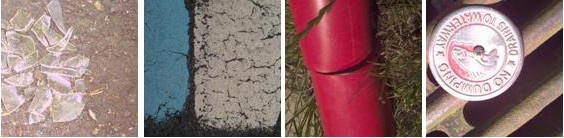}} \hspace{0.2cm} \subfigure[GTOS-Mobile]{\includegraphics[width=0.3\linewidth]{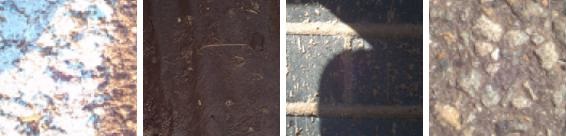}}
    
  \caption{Some mixed samples from training and test sets from each public texture recognition benchmark used in this work.\label{fig:datasets}}
\end{figure}

This work considers nine texture datasets to cover a variety of texture recognition tasks, including material classification, texture instance identification, and evaluating robustness to various texture transformations. 
The datasets include samples ranging from pure texture images captured in controlled environments, where the textures cover the whole image (usually depicting material surfaces or regular textures), to more complex images sourced from the internet (a.k.a. in-the-wild textures), which exhibit significant variability, including multiple objects, backgrounds, and overlaps. Some samples from these datasets are shown in Figure~\ref{fig:datasets}.

The evaluation protocols, including the training and test splits, differ across the datasets we use. They range in size from small scale, with around a thousand images, to large scale up to 100 thousand. Some adopt a single training/test split, while others provide pre-defined cross-validation splits. These details are shown in Table~\ref{tab:datasets} for each dataset.

\begin{table}[!htb]
	\centering
	\caption{\label{tab:datasets}Summary of the nine texture recognition datasets used in this work.}
	\resizebox{\linewidth}{!}{
	\begin{tabular}{|c|c|c|c|c|c|}
		\hline
            dataset & classes & train. & test & evaluation& description\\
            \hline
            Outex10 \cite{outex}& 24 & 480 & 3840 & single split & rotated grayscale textures\\
            Outex12 \cite{outex}& 24 & 480 & 4320 & single split & grayscale textures with varying illumination\\
            Outex13 \cite{outex}& 68 & 680 & 680 & single split & color textures\\
            Outex14 \cite{outex}& 68 & 680 & 1360 & single split & color textures with varying illumination\\

            DTD \cite{cimpoi2014describing}& 47 & 3760 & 1880 & 10-fold & texture attributes in-the-wild\\
            FMD \cite{sharan2010}& 10 & 900 & 100 & random 10-fold& material textures in-the-wild\\
            KTH-2-b \cite{caputo2005}& 11 & 3564 & 1188 & 4-fold& materials with varying scale, pose, and illumination\\

            GTOS \cite{xue2020}& 40 & 27284 & 6821 & 5-fold & outdoor ground materials \\
            
            GTOS-Mobile \cite{DEPNet} & 31 & 93945 & 6066 & single split & outdoor ground materials (mobile phone camera) \\
            
            \hline
	\end{tabular}
 }
\end{table}


In our tests, we evaluate a range of ViT backbones to analyze their performance on texture recognition tasks with the VORTEX approach.
For simplicity, each backbone is listed and described in Table~\ref{tab:backbones}, which details their pre-training, computational costs, and the number of features (hidden dimension $d$).
For better presentation, we omit repetitive citations to the ImageNet (IN) dataset~\cite{deng2009imagenet}. Instead, we refer to its two variants, featuring 1,000 and 21,000 classes, as IN-1k and IN-21k, respectively. For some backbones, we use both their IN-1k and IN-21k versions.


\begin{table}[!htb]
	\centering
	\caption{\label{tab:backbones}Details of each ViT backbone explored in this work with the proposed VORTEX method.}
	\begin{tabular}{|c|c|cc|c|}
		\hline
		& & \multicolumn{2}{c|}{feature extraction cost}& features\\
    	backbone &pre-training & GFLOPs & param. (M) & $d$ \\

            \hline
            ViT-S/16~\cite{steiner2021train}&IN-1k&4.2&21.7&384\\
            DeiT3-S/16~\cite{touvron2022deit}&IN-1k or IN-21k&4.2&21.7&384\\
            \hline
            ViT-B/16~\cite{dosovitskiy2020image}&IN-21k&16.9&85.8&768\\
            DeiT3-B/16~\cite{touvron2022deit}&IN-1k or IN-21k&16.9&85.8&768\\
            BEiTv2-B/16~\cite{peng2022beit}&IN-1k or IN-21k&12.7&85.8&768\\
            ViTamin-B~\cite{chen2024vitamin}&DataComp-1B~\cite{gadre2024datacomp}&21.8&87.1&768\\
            \hline
            DeiT3-L/16~\cite{touvron2022deit}&IN-1k or IN-21k&59.7&303.3&1024\\
            BEiTv2-L/16~\cite{peng2022beit}&IN-1k or IN-21k&44.8&303.4&1024\\
            ViTamin-L~\cite{chen2024vitamin}&DataComp-1B~\cite{gadre2024datacomp}&72.5&332.5&1024\\
            \hline
            ViTamin-XL~\cite{chen2024vitamin}&DataComp-1B~\cite{gadre2024datacomp}&95.8&434.7&1152\\
            ViT-H/14~\cite{ilharco2021openclip}&LAION-2B~\cite{schuhmann2022laionb} (OpenCLIP)&161.9&630.8&1280\\

        \hline
	\end{tabular}
\end{table}

We train classifiers, without tuning any hyperparameters, to assess the image representations extracted by ViT backbones and the VORTEX approach. These classifiers, which have been widely studied for decades, require significantly less training data compared to deep learning models. They are selected since some of the employed texture benchmarks have a small training set. Moreover, we consider this approach since our focus in this work is on the extracted features and their quality for texture analysis, rather than large-scale training and performance maximization. The following supervised classifiers are used:

\begin{itemize}
    \item \textbf{KNN}: $k$-Nearest Neighbors (k-NN), with $k=1$, is a simple yet effective method based on the Euclidean distance between samples, directly measuring the quality of the computed features.
    
    \item \textbf{LDA}: Linear Discriminant Analysis \cite{ripley2007pattern} is a linear decision classifier that models class-conditional densities from the data and applies the Bayes' rule to determine decision boundaries. We adopt the least-squares solver with automatic shrinkage using the Ledoit--Wolf lemma;

    \item \textbf{SVM}:  Support Vector Machine (SVM) algorithm \cite{cortes1995support} is a widely used method that leverages hyperplanes in a multidimensional feature space to achieve optimal separation between classes. Our experiments are performed with a linear kernel and $C=1$. 
    
\end{itemize}

\subsection{VORTEX ablations}

Firstly, we evaluate several aspects of the proposed VORTEX method. The results are presented and discussed in the following.

\subsubsection{Encoder Soup Size}

\begin{figure}[!htb]
    \centering
    \subfigure[]{\includegraphics[width=0.47\linewidth]{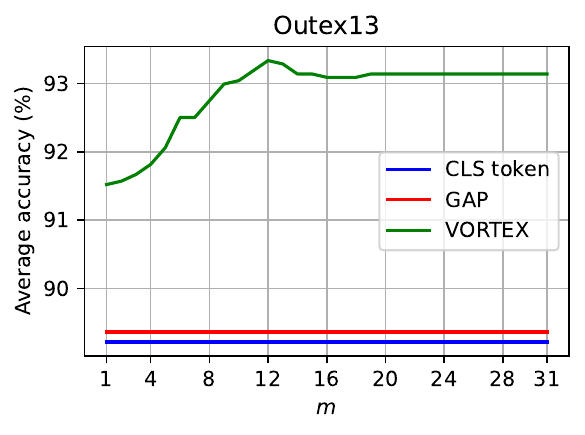}} \ \ \ \ \ \subfigure[]{\includegraphics[width=0.47\linewidth]{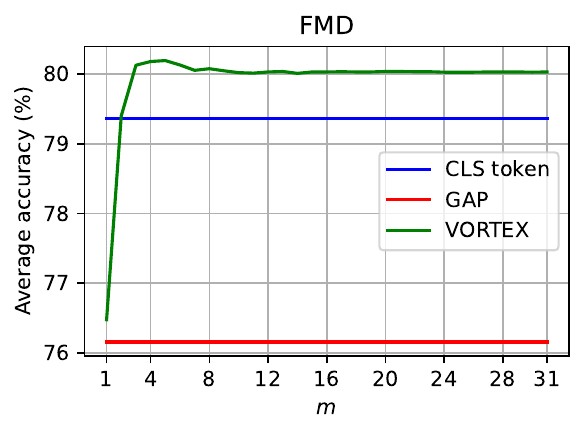}}
    
  \caption{Impacts of increasing the parameter $m$ (encoder soup size) of VORTEX compared to other feature extraction approaches (CLS token and GAP) when using the ViT-B/16 backbone (IN-21k). The result for each texture dataset (a, b) is the average classification accuracy among KNN, LDA, and SVM.\label{fig:soup}}
\end{figure}

The only parameter of VORTEX is the number of RAEs in the token encoding soup, $m$. We evaluate the use of a single RAE, as well as increasing it up to 31, and the results are shown in Figure~\ref{fig:soup}. For this experiment, we consider three different classifiers, KNN, LDA, and SVM. VORTEX is used with each one separately, and we report their average accuracy on each dataset (Outex13 and FMD). These datasets are selected to analyze two scenarios: pure textures (Outex13) and texture images from the internet, a.k.a. textures in the wild (FMD), measuring the impacts of $m$ in each case. The backbone is the vanilla ViT-B/16~\cite{dosovitskiy2020image} with IN-21k pre-training, which has a hidden dimension of $d=768$. With this backbone we compare three approaches (each one yields 768 features): the CLS token of the model as the final image representation; the GAP of the spatial tokens; and the proposed VORTEX representation.

The results show that VORTEX improves significantly with $m > 1$. We select $m=16$ for the next experiments, which we observed as a stable point in terms of performance on different texture recognition tasks. Compared to the other feature extraction approaches, VORTEX performs better for any choice of $m$ on the Outex13 dataset and requires $m>2$ for FMD. The gains of using VORTEX can be expressive compared to the CLS token and GAP, and our method consistently shows the best performance on both datasets. On the other hand, it is possible to notice that GAP was better than the CLS token on Outex13, while the opposite is observed on FMD, showing that the performance of these techniques may vary significantly.

\subsubsection{Classifier}\label{sec:classifiers}

We tested the proposed method using three classifiers: A simpler one (KNN), to highlight the quality of the features alone; and two more sophisticated approaches (LDA and SVM) for more challenging scenarios. The results (classification accuracy) are shown in Table \ref{tab:classifiers} for five datasets. We notice that SVM shows the highest average performance, followed closely by LDA, corroborating a stable performance of VORTEX features. Although KNN performs better for Outex10 and 12, simpler classification tasks, it falls behind on the more complex textures (Outex13 and 14 and FMD), which is to be expected. Therefore, we suggest SVM as the default classifier for VORTEX, since it is more likely to yield better performance on challenging tasks. We also use SVM for all the following experiments with VORTEX in this paper.

\begin{table}[!htb]
	\centering
	\caption{Texture classification accuracy of different classifiers (KNN, LDA, and SVM) using features from the ViT-B/16 (IN-21k) backbone coupled with the proposed VORTEX method. We use $m=16$, as previously defined.\label{tab:classifiers} 
 }

\begin{tabular}{|c|ccccc|c|}
\hline
 &Outex10&Outex12 &Outex13 & Outex14& FMD &Average \\
\hline 

KNN&96.9&95.7&90.1&75.3&71.2{\tiny$\pm$0.4}&85.9{\tiny$\pm$10.6}\\ 
LDA&94.0&92.9&95.0&77.6&84.7{\tiny$\pm$0.3}&88.9{\tiny$\pm$6.7}\\
SVM &95.1&94.0&94.1&77.9&84.2{\tiny$\pm$0.6}&89.0{\tiny$\pm$6.9} \\





 \hline 
 
	\end{tabular}
 
\end{table}

\subsubsection{Different Backbones and Sizes}

\begin{figure}[!htb]
    \centering
        \subfigure[Outex13]{\includegraphics[width=0.45\linewidth]{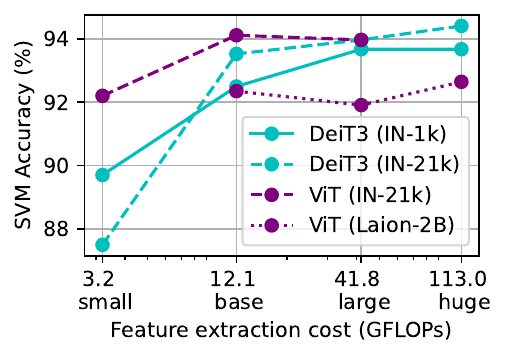}} \hspace{1cm} \subfigure[FMD]{\includegraphics[width=0.45\linewidth]{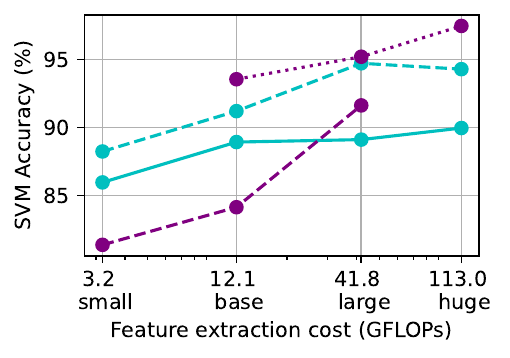}}
    
  \caption{Performance of the SVM classifier using VORTEX features with different ViT backbones by varying size and pre-training on two datasets (a-b).\label{fig:bkbsize}}
\end{figure}

To check the flexibility of VORTEX with different ViTs, we experiment with backbone variants from small to huge, and different pre-trainings (IN-1k, IN-21k, and LAION-2B). The results are shown in Figure~\ref{fig:bkbsize}, where the x-axis represents the backbone size in GFLOPs (after removing its original output head), and the y-axis is the SVM accuracy after training with VORTEX features from each backbone. We observe that VORTEX scales well with larger backbones and stronger pre-training, although the behavior may vary in different texture recognition scenarios. The larger backbones and pre-trainings stagnate on pure textures (Outex13, Figure~\ref{fig:bkbsize} (a)), where the images are more controlled. On the other hand, they are crucial for better performance on more complex scenarios, such as in-the-wild textures (FMD, (Figure~\ref{fig:bkbsize} (b)). Moreover, ViT models proposed after the original work, such as DeiT3, also shine on in-the-wild textures, such as its small and base versions. However, the highest results are achieved with the ViT large and huge variants with LAION-2B, where the gains of larger-scale pre-training truly shine. Nevertheless, these results show that VORTEX works well for various backbones and reflects their expected behavior (considering their size and pre-training) on different tasks.

\subsection{Comparison to other methods}

Here, we compare VORTEX to different feature extraction methods, backbones, texture analysis tasks, and SOTA texture recognition architectures.

\subsubsection{Off-the-shelf Feature Extraction}

\begin{figure}[!htb]
    \centering
        \subfigure[ViT-B/16 (IN-21k)]{\includegraphics[width=0.49\linewidth]{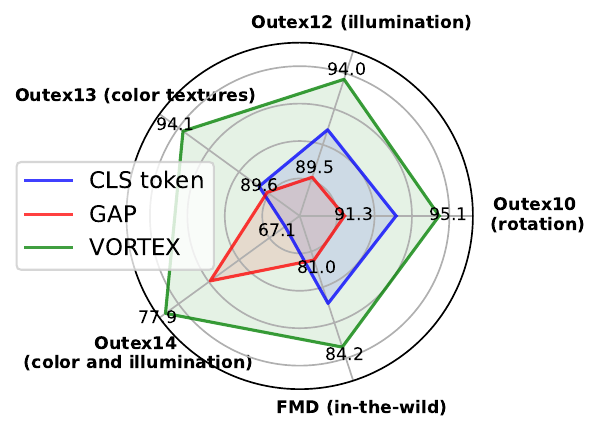}} \subfigure[ViT-B/16 (LAION-2B)]{\includegraphics[width=0.49\linewidth]{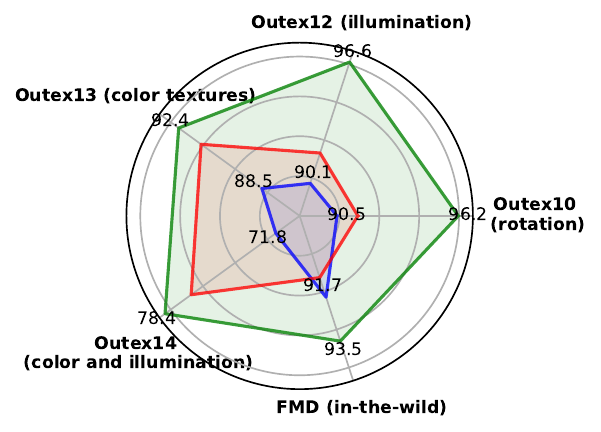}} \\

         \subfigure[ViT-L/16 (IN-21k)]{\includegraphics[width=0.49\linewidth]{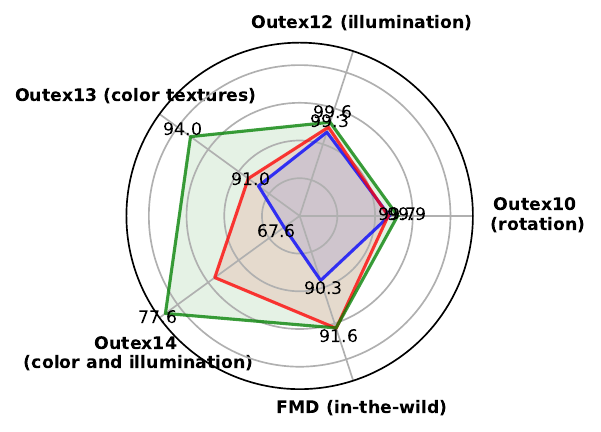}} \subfigure[ViT-L/14 (LAION-2B)]{\includegraphics[width=0.49\linewidth]{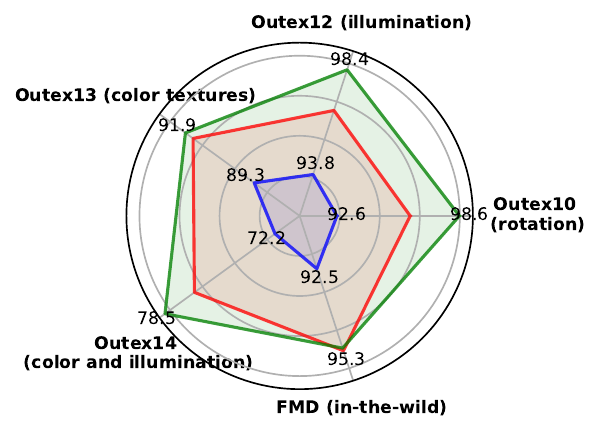}} \\

         \subfigure[ViT-H/14 (LAION-2B)]{\includegraphics[width=0.49\linewidth]{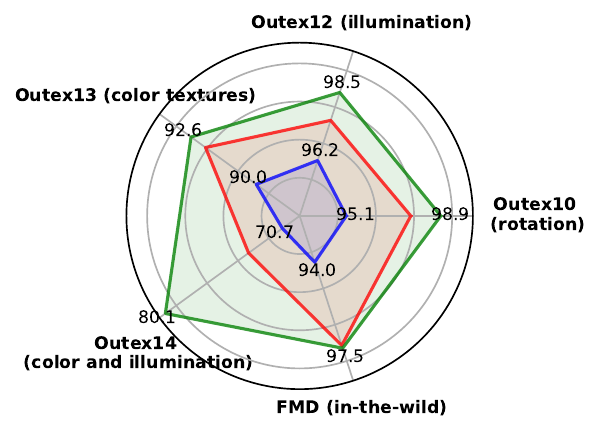}}
    
  \caption{Texture recognition skills, in terms of SVM classification accuracy, of different feature extraction methods (VORTEX, CLS token, and GAP) using various vanilla ViT backbones. \label{fig:poolingsvit}}
\end{figure}

The most common approach to performing off-the-shelf feature extraction with pre-trained ViTs is to extract the CLS token or the last vector used as input to the output linear head. Some works also employ the GAP of the spatial tokens to compose a single vector as the image representation. We compare VORTEX to these two approaches in a variety of texture recognition scenarios and using various ViT backbones. Figure~\ref{fig:poolingsvit} shows the obtained results (SVM classification accuracy) using the vanilla ViT architecture~\cite{dosovitskiy2020image} with either the original IN-21k pre-training or using LAION-2B. The results are shown as pentagons, where each vertex represents the performance of the method in a different texture recognition task. Therefore, we measure the ability of the methods to deal with texture rotation (Outex10), illumination changes (Outex12), color textures (Outex13), color textures with illumination changes (Outex14), and material textures in-the-wild (FMD).

\begin{figure}[!htb]
    \centering
        \subfigure[DeiT3-B/16 (IN-1k)]{\includegraphics[width=0.49\linewidth]{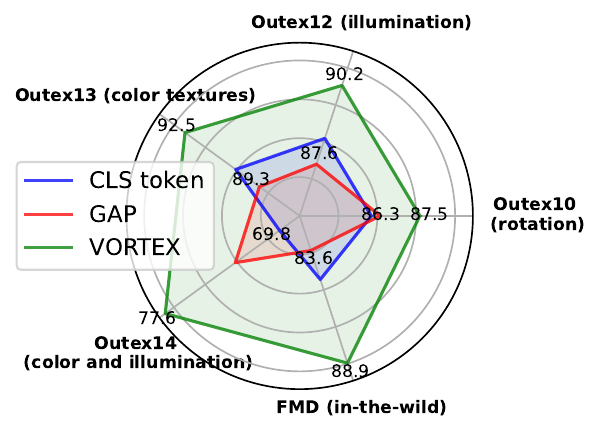}} \subfigure[DeiT-B/16 (IN-21k)]{\includegraphics[width=0.49\linewidth]{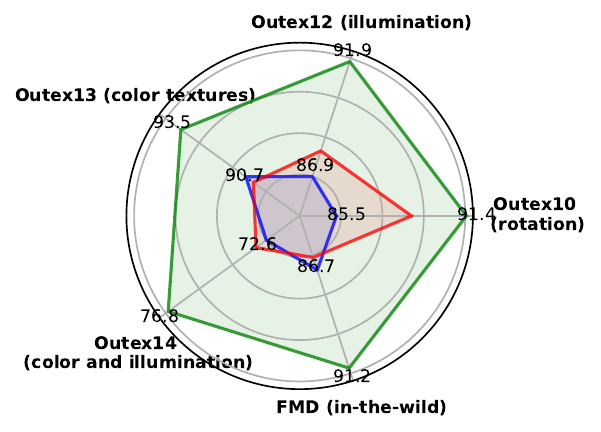}} \\
        \subfigure[DeiT3-L/16 (IN-1k)]{\includegraphics[width=0.49\linewidth]{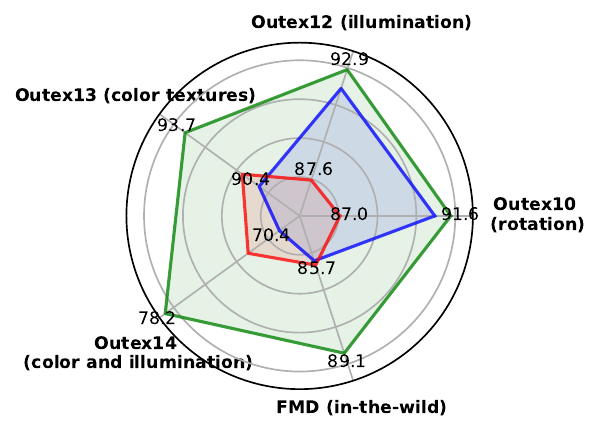}} \subfigure[DeiT-L/16 (IN-21k)]{\includegraphics[width=0.49\linewidth]{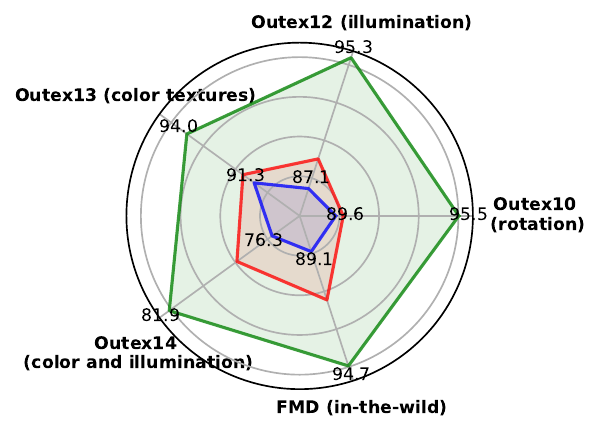}} \\
        \subfigure[DeiT3-H/14 (IN-1k)]{\includegraphics[width=0.49\linewidth]{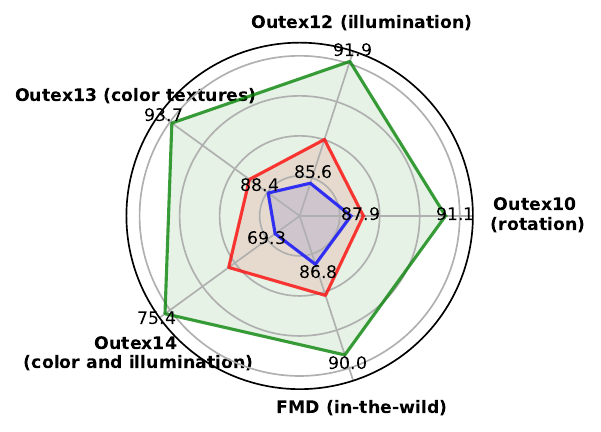}}       \subfigure[DeiT3-H/14 (IN-21k)]{\includegraphics[width=0.49\linewidth]{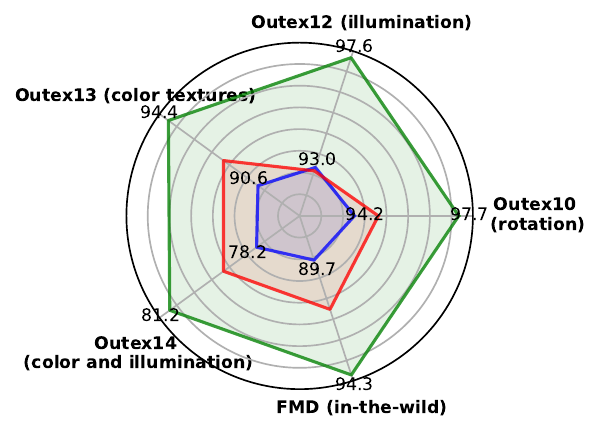}}

  \caption{Texture recognition skills, in terms of SVM classification accuracy, of different feature extraction methods (VORTEX, CLS token, and GAP) using various DeiT3 backbones.\label{fig:poolingsdeit}}
\end{figure}

The results with the vanilla ViT architecture shown in Figure~\ref{fig:poolingsvit} highlight the expressive gains obtained when using VORTEX while using the same number of features as the other methods. In most cases, the proposed method significantly improves the performance of the ViT backbones of different sizes and pre-trainings. In some cases, VORTEX keeps at least a similar performance compared to the best method between the CLS token or GAP. These methods, on the other hand, show an unstable performance where sometimes the CLS token is better, usually for ViT-B/16, and GAP surpasses it in most other cases. The gains when using VORTEX are especially higher for the ViT-B/16 architecture, closing the performance gap of this smaller variant compared to its larger counterparts. We also perform this analysis using DeiT3 backbones, and the results are shown in Figure~\ref{fig:poolingsdeit}. While, in this case, the same oscillating performance is observed for the CLS token and GAP, VORTEX shows even higher gains compared to the vanilla ViT backbones. VORTEX exhibits a stable and superior performance in general, posing as a better overall choice for feature extraction. These results corroborate the benefits of the VORTEX features and its good synergy with different and improved ViT backbones.

\subsubsection{Texture Recognition Methods}

In the following, we compare VORTEX to SOTA methods specifically designed for texture recognition, and some more general foundation CV models. Firstly, we consider methods using neural network backbones with IN-1k pre-training only, since they are more common in the literature. The results are shown in Table~\ref{tab:final-results} and represent the obtained classification accuracy and standard deviation (when pertinent) on five texture recognition tasks (DTD, FMD, KTH, GTOS, and GTOS-M). It is important to stress that most of the recent texture recognition methods in this category are designed over CNN backbones, especially ResNet50. This highlights a literature gap, where there are only a handful of texture analysis methods designed for and proven to work on more modern and larger backbones. Therefore, we also include the RADAM method, which explores backbones such as larger ResNets and ConvNeXts. The comparisons are divided according to backbone size. In the first group, which includes ResNet50 backbones and ViT-S, we observe that, in general, VORTEX is not able to surpass the CNN-based methods. However, the proposed method with DeiT3-S/16 achieves a new SOTA performance on GTOS-M (88.2\%), which is the largest dataset explored in this paper. This behavior highlights some known strengths and limitations of ViTs itself, such as the reliance on larger architectures and/or larger training sets. Therefore, using VORTEX with small ViTs and IN-1k-only pre-training is not the best approach for small datasets.


\begin{table}[!htb]
	\centering
	\caption{\label{tab:final-results}Classification results of several methods using backbones pre-trained on IN-1k only. All methods use a 224$\times$224 input size, except for EfficientNet-B5 which uses 512$\times$512 (indicated in parenthesis). The best results with methods from the literature are highlighted in \textbf{bold type}, while our results surpassing them are highlighted in \textcolor{red}{\textbf{red}}.
    }
        \rowcolors{2}{gray!20}{white}
	\begin{tabular}{lc|ccccc}
		\hline
		
    	method & backbone & DTD & FMD & KTH-2-b & GTOS & GTOS-M\\
        \hline  \hline

        
        


        DSRNet~\cite{zhai2020dsrnet} &ResNet50& \textbf{77.6}{\tiny$\pm$0.6} &86.0{\tiny$\pm$0.8}&85.9{\tiny$\pm$1.3}&85.3{\tiny$\pm$2.0}& \textbf{87.0} \\
        CLASSNet~\cite{chen2021classnet} &ResNet50& 74.0{\tiny$\pm$0.5} &86.2{\tiny$\pm$0.9}&87.7{\tiny$\pm$1.3}& \textbf{85.6}{\tiny$\pm$2.2}& 85.7 \\
        DFAEN~\cite{yang2022} &ResNet50& 73.2 &86.9& 86.3 & &86.9 \\
        RADAM~\cite{scabini2023radam} &ResNet50&75.6{\tiny$\pm$1.1}&85.3{\tiny$\pm$0.4}&88.5{\tiny$\pm$3.2}&81.8{\tiny$\pm$1.1}&81.0\\ 
        
        RADAM~\cite{scabini2023radam} &ConvNeXt-T&77.0{\tiny$\pm$0.7}&\textbf{88.7}{\tiny$\pm$0.4}&\textbf{90.7}{\tiny$\pm$4.0}&84.2{\tiny$\pm$1.7}&85.3\\  
        DBTrans \cite{liu2024dual}&ResNet50& 74.6& 87.5 & 88.4&&\\

HRNet \cite{qiu2024hrnet}&ResNet50& 73.8{\tiny$\pm$0.1}& 86.9{\tiny$\pm$0.9} & &&\\

HRNet \cite{qiu2024hrnet}&ResNeXt50& 74.5{\tiny$\pm$0.6}& 87.2{\tiny$\pm$1.2} & &&\\

         VORTEX &ViT-S/16&72.4{\tiny$\pm$0.9}&82.6{\tiny$\pm$0.6}&82.3{\tiny$\pm$5.9}&84.2{\tiny$\pm$1.5}&83.4\\ 

         VORTEX &DeiT3-S/16&74.2{\tiny$\pm$0.8}&86.0{\tiny$\pm$0.5}&88.9{\tiny$\pm$2.9}&84.1{\tiny$\pm$1.6}&\textcolor{red}{\textbf{88.2}}\\ 

         \hline
        DFAEN~\cite{yang2022} &Densenet161&76.1 &87.6& 86.6 & &\textbf{86.9}\\
        Multilayer-FV~\cite{lyra2022multilayerfv} &EfficientNet-B5 (512)&\textbf{78.9}& 88.7 &82.9 &&  \\


    RADAM~\cite{scabini2023radam}  &ConvNeXt-B&76.4{\tiny$\pm$0.9}&90.2{\tiny$\pm$0.2}&87.7{\tiny$\pm$5.6}&\textbf{84.1}{\tiny$\pm$1.6}&82.2\\    
    RADAM~\cite{scabini2023radam}  &Densenet161&75.0{\tiny$\pm$0.9}&84.9{\tiny$\pm$0.5}&88.6{\tiny$\pm$4.2}&83.3{\tiny$\pm$1.4}&85.8\\


    
    DBTrans \cite{liu2024dual}&DenseNet161& 76.9&\textbf{ 90.5}& 90.9&&\\ 
    HRNet \cite{qiu2024hrnet}&DenseNet161& 75.1{\tiny$\pm$0.8}& 86.7{\tiny$\pm$1.7} & &&\\ 
 
    Fractal pooling \cite{florindo2024fractal}&DenseNet161& & 89.3& \textbf{91.2}&&\\ 
    
     VORTEX &ViT-B/16&73.5{\tiny$\pm$0.7}&85.0{\tiny$\pm$0.5}&83.2{\tiny$\pm$6.4}&\textcolor{red}{\textbf{84.2}}{\tiny$\pm$2.4}&84.1\\

    VORTEX &DeiT3-B/16&76.6{\tiny$\pm$0.7}&88.9{\tiny$\pm$0.4}&90.0{\tiny$\pm$2.6}&\textcolor{red}{\textbf{85.0}}{\tiny$\pm$1.5}&\textcolor{red}{\textbf{88.0}}\\ 

    VORTEX &BEiTv2-B/16&\textcolor{red}{\textbf{79.9}}{\tiny$\pm$0.8}&\textcolor{red}{\textbf{93.4}}{\tiny$\pm$0.3}&\textcolor{red}{\textbf{93.1}}{\tiny$\pm$2.8}&\textcolor{red}{\textbf{86.1}}{\tiny$\pm$1.8}&\textcolor{red}{\textbf{88.0}}\\

    \hline

    RankGP-3M-CNN++~\cite{condori2021rankgp3mcnn} & (3 backbones) & & 86.2{\tiny$\pm$1.4} & \textbf{91.1}{\tiny$\pm$4.5} && \\

    RADAM~\cite{scabini2023radam}&ResNet152&72.1{\tiny$\pm$0.9}&83.0{\tiny$\pm$0.6}&87.7{\tiny$\pm$2.9}&82.2{\tiny$\pm$1.7}&82.9\\ 

    RADAM~\cite{scabini2023radam}&ConvNeXt-L&\textbf{77.4}{\tiny$\pm$1.1}&\textbf{89.3}{\tiny$\pm$0.3}&89.3{\tiny$\pm$3.4}&\textbf{84.0}{\tiny$\pm$1.8}&\textbf{85.8}\\

      VORTEX &DeiT3-L/16&\textcolor{red}{\textbf{77.9}}{\tiny$\pm$0.7}&89.1{\tiny$\pm$0.4}&91.0{\tiny$\pm$4.3}&\textcolor{red}{\textbf{85.3}}{\tiny$\pm$1.0}&\textcolor{red}{\textbf{91.7}}\\ 

        VORTEX &BEiTv2-L/16&\textcolor{red}{\textbf{82.5}}{\tiny$\pm$0.8}&\textcolor{red}{\textbf{93.9}}{\tiny$\pm$0.3}&\textcolor{red}{\textbf{91.2}}{\tiny$\pm$3.1}&\textcolor{red}{\textbf{88.2}}{\tiny$\pm$2.0}&\textcolor{red}{\textbf{88.2}}\\
        
      
\hline

	\end{tabular}
\end{table}

When bigger backbones are used, such as base and large ViT variants, VORTEX surpasses all the compared methods on all texture recognition tasks. These results highlight both the strengths of larger ViTs and the potential of VORTEX to deliver expressive gains compared to CNN-based texture recognition methods with comparable backbones. The proposed method with the base and large ViTs delivers uniformly strong results across diverse benchmarks, demonstrating robustness to dataset shifts and variations in domain characteristics. VORTEX with BEiTv2 backbones works particularly well in this scenario. For instance, there is a significant performance increase between RADAM with ConvNeXt-L (77.4{\tiny$\pm$1.1}) to VORTEX with BEiTv2-L (82.5{\tiny$\pm$0.8}) on the DTD dataset, the most challenging texture recognition benchmark considered here. VORTEX also shows competitive performance when using the DeiT3 backbones and achieves the highest result on GTOS-M (91.7\%) with DeiT3/L.

The next experiment, shown in Table~\ref{tab:final-results2}, includes comparisons with methods using backbones with pre-trainings stronger than IN-1k, ranging from IN-21K and LAION-2B to large-scale curated datasets like DataComp-1B and WIT 400m. In this case, we compare with the best CNN-based texture recognition method, RADAM, since it works with ConvNeXts with different pre-trainings, and we also include SOTA ViTs with published results on the DTD dataset. As previously discussed, RADAM is designed to work well in conjunction with larger CNN backbones, and this method achieves the best performance in most cases among other literature methods. Nevertheless, VORTEX surpasses RADAM in most cases except KTH-2-b with the large/x-large backbones and GTOS with the x-large backbones.
In general, VORTEX still delivers SOTA performance up to XL/huge backbone variants in most cases, showing its adaptability to larger architectures. Compared to other ViTs at the giant/huge variants and LAION pre-training, on DTD, VORTEX shows the best performance but falls short compared to RADAM with ConvNeXt-XXL. However, when using a smaller backbone such as the ViTamin-XL (efficiency is discussed in more depth later on), VORTEX surpasses RADAM, achieving the highest classification accuracy reported to date (87.6{\tiny$\pm$0.7}) on the challenging DTD benchmark. The proposed method with ViTamin-L/XL also sets a new SOTA on FMD and GTOS-M.

\begin{table}[!htb]
	\centering
	\caption{\label{tab:final-results2}Classification results of several methods using backbones pre-trained with larger datasets. All backbones use a 224$\times$224 input size, except for the cases indicated in parenthesis (336 or 384). The best results with methods from the literature are highlighted in \textbf{bold type}, while our results surpassing them are highlighted in \textcolor{red}{\textbf{red}.}
    }
        \rowcolors{2}{gray!20}{white}
	\begin{tabular}{lcc|ccccc}
		\hline
		
    	method & backbone & pre-training& DTD & FMD & KTH-2-b & GTOS & GTOS-M\\
        \hline  \hline





                fine-tuning~\cite{zhang2022bamboo} & ViT-B/16 &Bamboo 69m &81.2 &&&&\\ 

        
        RADAM~\cite{scabini2023radam} &ConvNeXt-B &IN-21K&82.8{\tiny$\pm$0.9}&\textbf{94.0}{\tiny$\pm$0.2}&91.8{\tiny$\pm$4.1}&86.6{\tiny$\pm$1.7}&\textbf{87.1}\\ 
        
RADAM &ConvNeXt-B&LAION-2B&83.2{\tiny$\pm$0.7}&93.6{\tiny$\pm$0.3}&\textbf{92.4}{\tiny$\pm$1.8}&\textbf{86.9}{\tiny$\pm$1.8}&86.6\\
MLCD~\cite{an2025multi} & ViT-B/32 &COYO-700M & \textbf{83.5} &&&&\\

 
 VORTEX &DeiT3-B/16&IN-21k&79.2{\tiny$\pm$1.0}&91.2{\tiny$\pm$0.3}&\textcolor{red}{\textbf{92.6}}{\tiny$\pm$3.9}&85.8{\tiny$\pm$1.6}&\textcolor{red}{\textbf{88.1}}\\ 

 VORTEX &BEiTv2-B/16&IN-21k&82.2{\tiny$\pm$0.9}&\textcolor{red}{\textbf{94.9}}{\tiny$\pm$0.2}&\textcolor{red}{\textbf{94.6}}{\tiny$\pm$3.6}&\textcolor{red}{\textbf{87.1}}{\tiny$\pm$1.8}&\textcolor{red}{\textbf{87.3}}\\ 
 VORTEX &ViTamin-B&DataComp-1B&\textcolor{red}{\textbf{83.6}}{\tiny$\pm$0.8}&\textcolor{red}{\textbf{95.7}}{\tiny$\pm$0.3}&89.0{\tiny$\pm$4.6}&86.3{\tiny$\pm$1.1}&85.9\\ 

 \hline

        CLIP~\cite{radford2021clip} & ViT-L/14 (336)& WIT 400m & 83.0 &&&&\\        
        $\mu$2Net+~\cite{gesmundo2022continual} & ViT-L/16 (384) & IN-21K& 82.2 &&&&\\
 RADAM~\cite{scabini2023radam} &ConvNeXt-L& IN-21K&84.0{\tiny$\pm$1.0}&\textbf{95.2}{\tiny$\pm$0.4}&91.3{\tiny$\pm$4.1}&85.9{\tiny$\pm$1.6}&87.3\\

RADAM &ConvNeXt-L& LAION-2B&83.7{\tiny$\pm$0.6}&94.5{\tiny$\pm$0.2}&\textbf{96.4}{\tiny$\pm$1.9}&\textbf{86.1}{\tiny$\pm$2.1}&\textbf{87.6}\\ 

 MLCD~\cite{an2025multi} & ViT-L/14& LAION-400M & \textbf{84.6} &&&&\\
 
 
 VORTEX &DeiT3-L/16&IN-21k&81.8{\tiny$\pm$0.9}&94.7{\tiny$\pm$0.3}&92.5{\tiny$\pm$4.7}&\textcolor{red}{\textbf{86.4}}{\tiny$\pm$2.0}&83.8\\ 
 VORTEX &BEiTv2-L/16&IN-21k&\textcolor{red}{\textbf{84.7}}{\tiny$\pm$0.7}&\textcolor{red}{\textbf{96.9}}{\tiny$\pm$0.3}&94.0{\tiny$\pm$2.2}&\textcolor{red}{\textbf{87.6}}{\tiny$\pm$2.6}&\textcolor{red}{\textbf{90.7}}\\ 
 VORTEX &ViTamin-L&DataComp-1B clip&\textcolor{red}{\textbf{87.1}}{\tiny$\pm$0.6}&\textcolor{red}{\textbf{98.8}}{\tiny$\pm$0.1}&95.2{\tiny$\pm$1.1}&\textcolor{red}{\textbf{88.8}}{\tiny$\pm$1.2}&\textcolor{red}{\textbf{91.8}}\\

 \hline 
         RADAM~\cite{scabini2023radam} &ConvNeXt-XL& IN-21K&83.7{\tiny$\pm$0.9}&95.2{\tiny$\pm$0.3}&94.4{\tiny$\pm$3.8}&87.2{\tiny$\pm$1.9}&90.2\\ 

RADAM &ConvNeXt-XXL&LAION-2B&\textbf{86.3}{\tiny$\pm$0.8}&\textbf{97.1}{\tiny$\pm$0.1}&\textbf{97.4}{\tiny$\pm$1.9}&\textbf{89.4}{\tiny$\pm$1.3}&\textbf{91.5}\\ 

OpenCLIP \cite{ilharco2021openclip} & ViT-G/14& LAION-2B& 86.0 &&&&\\
DINOv2 \cite{oquab2023dinov2} & ViT-g/14 & LVD-142M& 84.6 &&&&\\


 VORTEX &ViT-H/14&LAION-2B&85.7{\tiny$\pm$0.7}&\textcolor{red}{\textbf{97.4}}{\tiny$\pm$0.2}&93.1{\tiny$\pm$1.9}&87.1{\tiny$\pm$2.4}&89.1\\ 
 VORTEX &ViTamin-XL&DataComp-1B&\textcolor{red}{\textbf{87.6}}{\tiny$\pm$0.7}&\textcolor{red}{\textbf{98.6}}{\tiny$\pm$0.2}&95.1{\tiny$\pm$2.8}&88.9{\tiny$\pm$1.8}&\textcolor{red}{\textbf{93.8}}\\ 

     \hline

	\end{tabular}
\end{table}

Although VORTEX outperforms most methods, RADAM shows strong competition in specific cases, indicating areas where VORTEX can be further optimized. For instance, the CNN architecture appears to be a better approach for pure textures with high variability and low noise, as demonstrated by the results on KTH-2-b and GTOS. On the other hand, VORTEX achieves the best performance on DTD, FMD, and GTOS-M, datasets with less image acquisition control and a larger variety of textures, highlighting the benefits of ViT features in these scenarios.

\subsection{Efficiency}

\begin{figure}[!htb]
    \centering
        \subfigure[IN-1k pre-training.]{\includegraphics[width=0.49\linewidth]{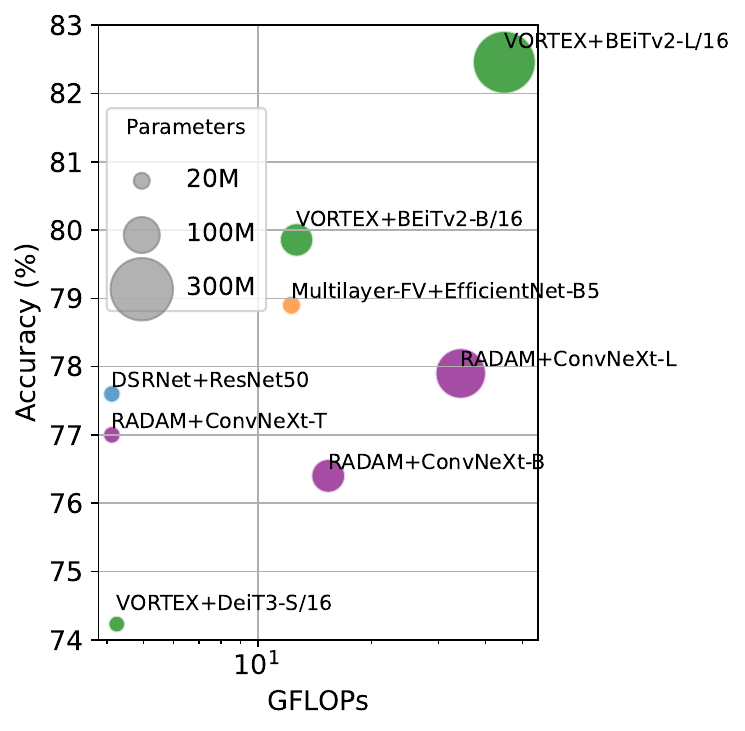}}  \subfigure[Other pre-trainings.]{\includegraphics[width=0.49\linewidth]{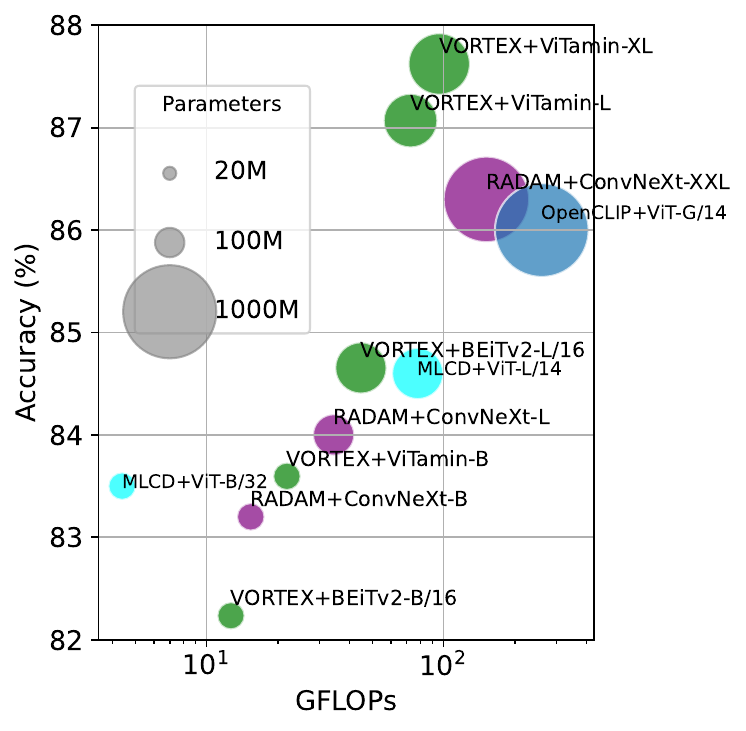}}

  \caption{Efficiency analysis on the DTD dataset, including the proposed VORTEX method and the best methods from the literature on this dataset. Cost is measured by GFLOPs and the number of parameters, and performance is the classification accuracy.\label{fig:efficiency}}
\end{figure}

To better illustrate the efficiency of VORTEX compared to other techniques, we measure the number of floating point operations, in billions (GFLOPs), and the number of parameters, in millions, of the backbones each method employs. We show these costs in function of classification accuracy on DTD, the most challenging dataset with more available results for comparison. The results are shown in Figure~\ref{fig:efficiency} divided between the best methods with IN-1k pre-training only (a), and those with other pre-trainings (b). For IN-1k pre-training, DSRNet with ResNet50 is the best low-cost alternative, while VORTEX with DeiT3-S/16, although having a similar cost, exhibits lower performance. With larger backbones, VORTEX with BEiTv2-L/16 achieves a considerably higher classification accuracy but has a larger number of parameters (303M vs 196M) and GFLOPs (44.8 vs 34.3) than RADAM with ConvNeXt-L. The Multilayer-FV method surpasses RADAM while having a much lower cost (28M parameters and 12.3 GFLOPs), highlighting the benefits of the EfficientNet-B5 architecture in this scenario. However, it is important to stress that Multilayer-FV requires a 512$\times$512 input size to reach its performance in this case, which is considerably higher than the 224$\times$224 input size of VORTEX. Therefore, VORTEX using BEiTv2-B/16 is a strong choice in terms of efficiency, showing the second highest accuracy with relatively moderate parameter size (85M) and GFLOPs (12.7) compared to Multilayer-FV. Moreover, VORTEX performance may increase when using ViTs with higher input-size resolution.

As previously discussed, VORTEX with ViTamin-XL achieves the highest accuracy, setting a new SOTA with large-scale pre-training on the DTD dataset. Compared to RADAM with ConvNeXt-XXL, the best competitor from the literature, VORTEX has significantly fewer parameters (434M vs 843M) and GFLOPs (95.8 vs 151.6), as shown in Figure~\ref{fig:efficiency} (b). It also surpasses the SOTA when using ViTamin-L, which has even fewer parameters (332M) and GFLOPs (72.5), making it a more computationally efficient option. VORTEX with BEiTv2-L/16 is a good intermediate choice, surpassing MLCD with ViT-L/14 and RADAM with ConvNeXt-L. VORTEX with ViTamin-B is a competitive alternative at a smaller cost. Although it has higher GFLOPs than MLCD using ViT-B/32 (21.8 vs 4.4), it achieves a 0.1\% higher accuracy and has a similar number of parameters.

Another key advantage of VORTEX, in terms of cost, is that it operates purely as a feature extractor coupled with a simple linear classifier (SVM), bypassing the need for fine-tuning the large ViT backbones while still achieving SOTA performance across diverse datasets. This approach, which is also employed by RADAM, significantly reduces computational costs and simplifies the training pipeline~\cite{scabini2023radam}. Focusing on pre-trained backbones to extract high-quality features, rather than optimizing training/fine-tuning, enables us to outperform many competitors, especially in scenarios with limited labeled data or constrained computational resources. However, while RADAM with ConvNeXt-XXL performs well in some scenarios, as shown before (Table~\ref{tab:final-results2}), its higher computational budget highlights the superiority of VORTEX in balancing performance and cost. Moreover, the performance and efficiency of VORTEX across different backbone sizes and pre-trainings make it a versatile solution for diverse computational environments.

\section{Conclusion}

In this work, we introduce VORTEX, a novel texture recognition method leveraging pre-trained ViTs as feature extractors without requiring fine-tuning. By aggregating multi-depth token embeddings and employing orderless randomized encodings over frozen ViTs, VORTEX captures hierarchical features that enhance texture representations fed to a linear SVM for training and inference. Extensive experiments across nine diverse datasets demonstrated its SOTA performance. VORTEX also proved highly adaptable, delivering robust results with various ViT backbones and pre-training strategies while maintaining computational efficiency. It consistently outperformed or matched competitive approaches, such as texture-specific ones like RADAM and other CNN-based methods, and general vision ViTs like DINOv2 and CLIP models.

Our work points to a significant shift in texture recognition, as VORTEX enables using modern foundation models based on transformers, which are continually emerging in CV. Therefore, VORTEX shows the potential to bridge the gap between CNN-based texture analysis and ViTs. Future work may focus on several directions to enhance VORTEX, including exploring its application to specific problems, such as medical imaging and remote sensing, where textures are crucial. Additionally, incorporating domain-specific pre-training strategies could improve its performance in specialized settings. Addressing backbones that deal with varying input sizes could also benefit the method by allowing it to process higher-resolution textures. This may also be achieved by investigating the integration of VORTEX with hybrid architectures, combining the strengths of ViTs and CNNs. Finally, as new foundation ViTs emerge, extending VORTEX to work seamlessly with these architectures will ensure its continued relevance and effectiveness.

\section*{Acknowledgements}

L. Scabini acknowledges funding from the São Paulo Research Foundation (FAPESP) (grants \#2023/10442-2 and \#2024/00530-4). 
K. M. Zielinski acknowledges support from FAPESP (grant \#2022/03668-1) and Higher Education Personnel Improvement Coordination (CAPES) (grant  \#88887.631085/2021-00).
R. T. Fares acknowledges support from FAPESP (grant \#2024/01744-8).
L. C. Ribas acknowledges support from FAPESP (grant \#2023/04583-2). 
O. M. Bruno acknowledges support from the Brazilian National Council for Scientific and Technological Development (CNPq) (grant \#305610/2022-8) and FAPESP (grants \#2018/22214-6 and \#2021/08325-2).

\bibliographystyle{unsrt}
\bibliography{arxiv.bib}

\end{document}